\documentclass[letterpaper, 10 pt, conference]{ieeeconf}  

\usepackage{booktabs}
\usepackage{color,colortbl}
\usepackage[colorlinks=true, allcolors=blue]{hyperref}
\usepackage{cite}
\usepackage{listings}
\usepackage{amsmath,amssymb,amsfonts}
\usepackage{algorithmic}
\usepackage{graphicx}
\usepackage{textcomp}
\usepackage{hyperref}
\usepackage{subcaption}
\usepackage{caption}
\usepackage{float}
\usepackage{longtable}
\usepackage{array}
\usepackage{xcolor}
\usepackage{xtab,afterpage} 
\usepackage{longtable}
\usepackage{multicol}

\IEEEoverridecommandlockouts                              

\title{\LARGE \bf
ML-SceGen: A Multi-level Scenario Generation
Framework}
\author{Yicheng Xiao, Yangyang Sun and Yicheng Lin}

\begin{document}

\maketitle

\begin{abstract}
Current scientific research witnesses various attempts at applying Large Language Models for scenario generation but is inclined only to comprehensive or dangerous scenarios. 
In this paper, we seek to build a three-stage framework that not only lets users regain controllability over the generated scenarios but also generates comprehensive scenarios containing danger factors in uncontrolled intersection settings. 
In the first stage, LLM agents will contribute to translating the key components of the description of the expected scenarios into Functional Scenarios. 
For the second stage, we use Answer Set Programming (ASP) solver Clingo to help us generate comprehensive logical traffic within intersections. 
During the last stage, we use LLM to update relevant parameters to increase the critical level of the concrete scenario.
\end{abstract}


\section{INTRODUCTION}
\label{sec:intro}

Over the years, the development of autonomous driving has soared. Society witnessed the gradual advent of driverless cars, as the notion no longer stays online but can be captured by our own eyes. However, as the technology of autonomous systems is evolving, from ontology-based methods to deep learning methods \cite{lan2022vision,lan2024dir}, it is always hard to generate scenarios that can ensure both comprehensiveness and critical levels. \textbf{Comprehensiveness} requires a thorough search in the action space while evaluating \textbf{critical levels} sometimes requires crash data that are somehow unpatterned. 

Current research in autonomous driving systems heavily relies on Machine Learning \cite{lan2022class,gao2021neat}, Deep Learning, or Reinforcement Learning methods, which leads to a lack of interpretability. 
The outcome usually fails to attain the expected controllability of the users. Also, these methods rely heavily on real-world data, which might be biased and expensive to achieve. 
Using simulators and generating scenarios within the simulators are becoming an affordable and effective alternate solution. 

Currently, there are three main types of simulators:
\begin{itemize}
    \item \textbf{Render Engine-Based Simulator}: These simulators, such as CARLA \cite{dosovitskiy2017carla}, require manual modeling of scenes and objects. However, they face two key challenges: a domain gap, as the simulated environments often differ from reality, and the difficulty of constructing realistic and complex scenes. The scenario realism is restricted by asset modeling and rendering qualities, and adding digital assets to the scene is difficult to get started due to the complexity of the simulation system.
    \item \textbf{Video Generation-Based Simulators}: These simulators utilize techniques like Diffusion Models to generate environments based on real-world data. While the results closely resemble reality, the process introduces significant randomness and lacks control, leading to occasional abrupt changes in the generated data. They struggle to maintain view consistency and face challenges in importing external digital assets due to the absence of 3D spatial modeling \cite{chatscene}. 
    \item \textbf{Scene Reconstruction-Based Simulators}: Examples like UniSim \cite{yang2023unisim} leverage neural rendering methods such as NeRF (Neural Radiance Fields) \cite{mildenhall2020nerf} to reconstruct realistic scenes. A notable limitation is that these simulators struggle to incorporate external digital assets.
\end{itemize}

From the above descriptions, we can discover the problems of the current generation methods in simulators: lack of controllability, bad scalability, and distance from reality \cite{lan2024sustechgan}. 
We therefore hope to devise a scalable framework that supports the generation of scenarios from Functional to Logical, from Logical to Concrete. Our contributions are as follows:
\begin{itemize}
    \item We identify \textbf{ASP solver} to be a core part of our logical scenario generation framework to achieve comprehensiveness.
    \item We leverage the ability of \textbf{Large Language Model} to help apprehend the possible danger factors and incorporate them in the generated scenarios to increase the critical levels. 
    \item We have conducted relevant experiments to demonstrate the controllability of our framework. 
\end{itemize}

\section{RELATED WORK}
\label{sec:related}

\subsection{Chatscene and TTSG}

\textbf{ChatScene} is a novel large language model (LLM)-based agent designed to generate safety-critical scenarios for autonomous vehicles. Its primary function is to translate textual descriptions of traffic scenarios into executable simulations within the CARLA simulation environment. The combination of textual generation, scenario decomposition, knowledge retrieval from a pre-constructed database, and execution of simulation scripts supports ChatScene’s effectiveness in generating safety-critical scenarios. However, the language ChatScene chose to generate is Scenic, which is a less popular language than OpenDrive, OpenScenario format, and a less controllable domain-specific language for it only supports a limited range of control actions. Also, the generation process of ChatScene is unguided, which leads to a high error rate and requires post-processing of the generated result. \cite{chatscene}

\textbf{Text-to-traffic scene generation (TTSG)} framework is to autonomously generate diverse and customizable traffic scenarios from natural language descriptions, enhancing the training of autonomous vehicles. The framework aims to overcome the limitations of traditional methods, which often restrict scenario diversity by relying on predefined paths and fixed conditions. Users can specify various parameters such as weather conditions, vehicle types, and road signals, while the system dynamically selects starting points and generates scenes from scratch, facilitating the representation of both critical and routine traffic situations. However, the problem with the method is that it cannot automatically generate comprehensive scenario under a given circumstances. \cite{ttsg}

\subsection{Specification, Validation, and Verification of Autonomous System}

Marius and Joseph presented a multilevel semantic framework designed for the specification and validation of Autonomous Driving Systems (ADS) in 2021. 
\cite{spec} It addresses the complexities of integrating these systems with non-explainable AI technologies by formalizing physical environments as directed metric graphs. 
The framework introduces three logics: Metric Configuration Logic (MCL) for mapping parameters, Mobile Metric Configuration Logic (M2CL) for representing dynamic object distributions, and Temporal M2CL for time-related properties \cite{lan2022time,lan2019evolving}. 
This systematic approach facilitates the creation of detailed traffic rules and dynamic scenarios while tackling validation challenges like runtime verification and satisfiability. For scenario generation, using the logical framework can help define logical scenarios and thus provide proper guidance towards concrete scenario generation. 

Li et al. presented a comprehensive framework for validating autonomous driving systems (ADS) by integrating an industrial simulator with a scenario generator and a runtime verification monitor based on the framework proposed above. \cite{simbase} The proposed methodology focuses on systematically generating scenarios that explore high-risk situations, particularly at complex junctions, to uncover defects that standard random exploration might miss.

\subsection{Answer Set Programming Solver}

Answer Set Programming (ASP) offers a simple and powerful modeling language to describe combinatorial problems as logic programs. The \textit{clingo} system then takes such a logic program and computes answer sets representing solutions to the given problem. Clingo is part of the Potassco project for Answer Set Programming (ASP), combining a grounder (Gringo) and a solver (Clasp) into a single tool. Problems are expressed in terms of \textbf{rules, facts, and constraints}, which specify the relationships and conditions to be satisfied. 
Solutions, called answer sets, represent \textbf{possible scenarios that satisfy all rules in the program}. 
Usually, after formulating the problems into \texttt{.lp} file, the grounder will ground (Gringo) the program to generate a propositional formula, and the solver (Clasp) will solve the formula to find answer sets that satisfy constraints. 
An example of a calling program is shown as follows.

\begin{lstlisting}[language=Python, caption=Example Hanoi Program using ASP]
% Facts
peg(a;b;c).
disk(1..6).
init_on(1..6,a).
goal_on(1..6,c).
moves(63).

% Rules
{ move(D,P,T) : disk(D), peg(P) } = 1 :- moves(M),  T = 1..M.

move(D,T)   :- move(D,_,T).
on(D,P,0)   :- init_on(D,P).
on(D,P,T)   :- move(D,P,T).
on(D,P,T+1) :- on(D,P,T), not move(D,T+1), not moves(T).
blocked(D-1,P,T+1) :- on(D,P,T), not moves(T).
blocked(D-1,P,T)   :- blocked(D,P,T), disk(D).

% Constraints
:- move(D,P,T), blocked(D-1,P,T).
:- move(D,T), on(D,P,T-1), blocked(D,P,T).
:- goal_on(D,P), not on(D,P,M), moves(M).
:- { on(D,P,T) } != 1, disk(D), moves(M), T = 1..M.

% Display
#show move/3.
\end{lstlisting}

\subsection{Scenario Generator (OpenScenario XML)}
 
OpenScenario and OpenDrive are standards proposed by the Association for Standardization of Automation and Measuring Systems (ASAM) and are widely used in autonomous driving and traffic simulation environments.

OpenScenario is an XML-based file format designed to describe dynamic traffic scenarios for testing and validation of autonomous driving systems \cite{lan2023virtual,xu2019online}. It specifies maneuvers, actions, and events involving vehicles, pedestrians, and other entities in a simulation. Key features include:
\begin{itemize}
    \item Definition of traffic participants and their actions.
    \item Support for complex event triggering mechanisms.
    \item Compatibility with traffic simulation tools and autonomous driving platforms.
\end{itemize}

For detailed specification, you can refer to \href{https://www.asam.net/index.php?eID=dumpFile&t=f&f=4092&token=d3b6a55e911b22179e3c0895fe2caae8f5492467}{OpenScenario User Guide.}

OpenDrive is a format used to define road networks and static environments in a simulation. It models lanes, road geometries, intersections, and traffic signs, making it a cornerstone for creating realistic road layouts. Key features include:
\begin{itemize}
    \item Representation of detailed road layouts and lane structures.
    \item Support for multiple coordinate systems and elevation profiles.
    \item Integration with OpenScenario for complete scenario modeling.
\end{itemize}

For detailed specification, you can refer to \href{https://publications.pages.asam.net/standards/ASAM_OpenDRIVE/ASAM_OpenDRIVE_Specification/latest/specification/index.html}{OpenDrive User Guide.}

Together, OpenScenario and OpenDrive enable the creation of rich and dynamic simulation environments, supporting the development and validation of autonomous driving systems and ADAS (Advanced Driver Assistance Systems).

\texttt{scenariogeneration} is a Python package that supports the generation of OpenScenario and OpenDrive format files. It contains two sub-packages, \texttt{xosc} and \texttt{xodr}, which function as XML generators. The \texttt{xosc} subpackage allows for the design of concrete scenarios that conform to OpenScenario V1.1.0, including the specification of the number of vehicles, their initial states, and their behaviors. Meanwhile, the \texttt{xodr} subpackage can generate the corresponding road networks and static environments.

For detailed specification, you can refer to \href{https://github.com/pyoscx/scenariogeneration}{repository of scenariogeneration.}

\subsection{Large Language Models}

\textbf{Large Language Models} (LLMs) are advanced artificial intelligence systems trained on vast amounts of textual data to understand and generate human-like language \cite{yi2024key}. 
Large language models have demonstrated remarkable abilities in understanding context, generating answers, and handling complex tasks. The innate common sense embedded within large language models can help narrow the gap between simulated and real-world environments \cite{yang2024llm4drivesurvey}, which makes LLMs a good choice to analyze the natural-language-described scenario and give proper suggestions for increasing emergency level for a scenario.

GPT \cite{brown2020languagemodels} is a pioneering work to generate human-like content, followed by the most popular LLMs, GPT-3.5 and GPT-4, which provide more intelligent capabilities like chatting and coding. Some other outstanding models like LLaMA \cite{touvron2023llama2} also perform well in dealing with language understanding tasks. 

Based on LLM, \cite{chatscene} and \cite{ttsg} use LLM to generate specific scenario codes using natural language descriptions. \cite{brown2020languagemodels} exploit the power of collaborative LLM agents to write Python files in simulation for autonomous driving \cite{lan2019simulated,lan2019evolutionary}. 
The idea of defining a group of well-organized agents to form operating procedures with conversation and code programming \cite{hong2024metagpt} is enlightening and practical with the help of tools like LangChain. 
But temporarily the code-writing ability of LLMs is unstable and error-prone due to the hallucination problem. This means the generated codes still need manual adjustment before being applied.

\section{METHODOLOGY}
\label{sec:methodology}
\subsection{Motivation}

From the above sections, we are striving to set up a framework where we can not only generate comprehensive scenarios through formulating the traffic and getting satisfiable solutions but also use LLM to analyze the critical factors and integrate them into the given scenario. Based on this, we propose the following three-stage framework. 

\begin{figure}
    \centering
    \includegraphics[width=1\linewidth]{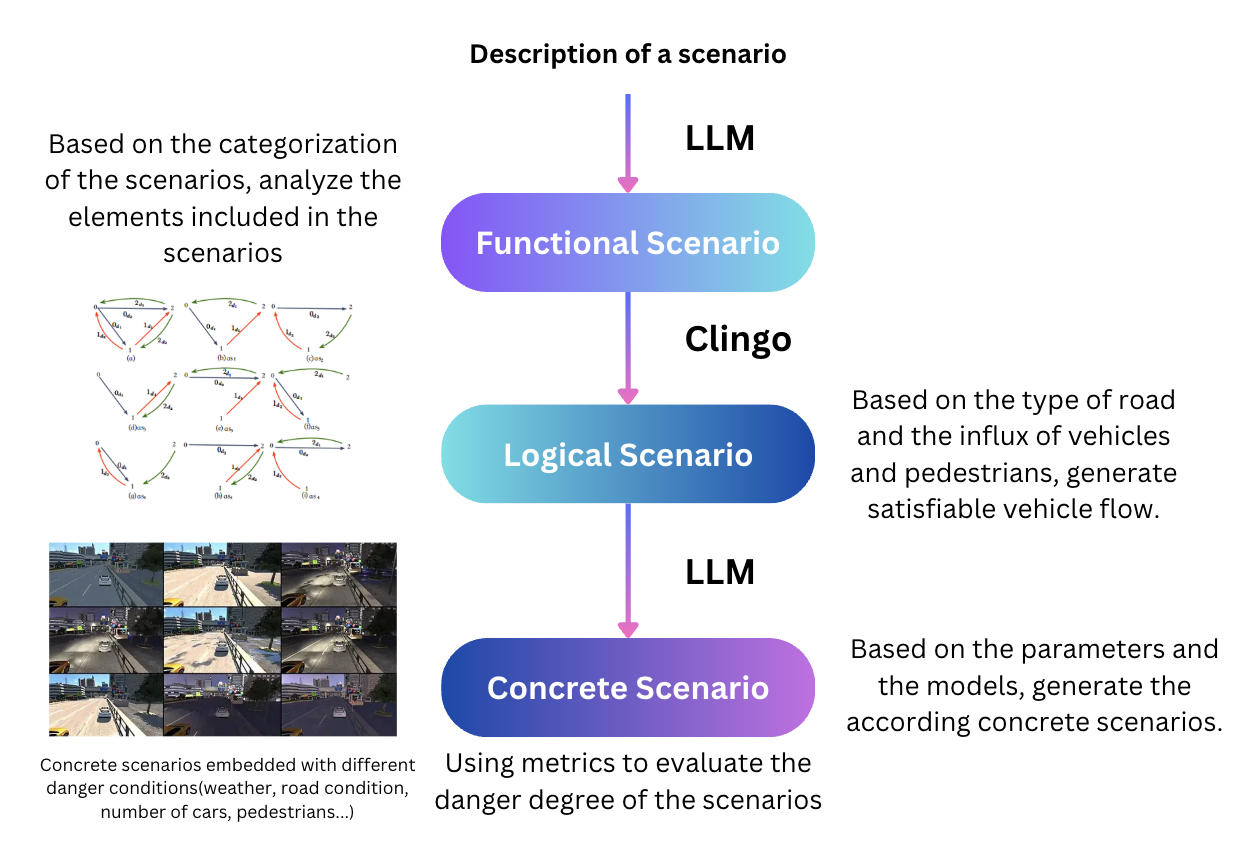}
    \caption{Three-stages framework for scenario generation}
    \label{fig:framework}
\end{figure}

\subsection{Framework Setup}

We have divided the generation framework into three stages, which will generate scenarios from functional scenarios to logical scenarios to concrete scenarios. The implementation detailed is as follows:

\subsubsection{Stage 1: Multi-agents Parser}

In this project, we leverage the collaboration of LLM agents to tackle our problem. 

For a description we prompt into the LLM agents, we are hoping to use LLM to extract and reason valuable information for us to use in the latter modules. The first-stage agents will separately parse the inputs for the second stage and the input for the third stage. 

\begin{figure}
    \centering
    \includegraphics[width=1\linewidth]{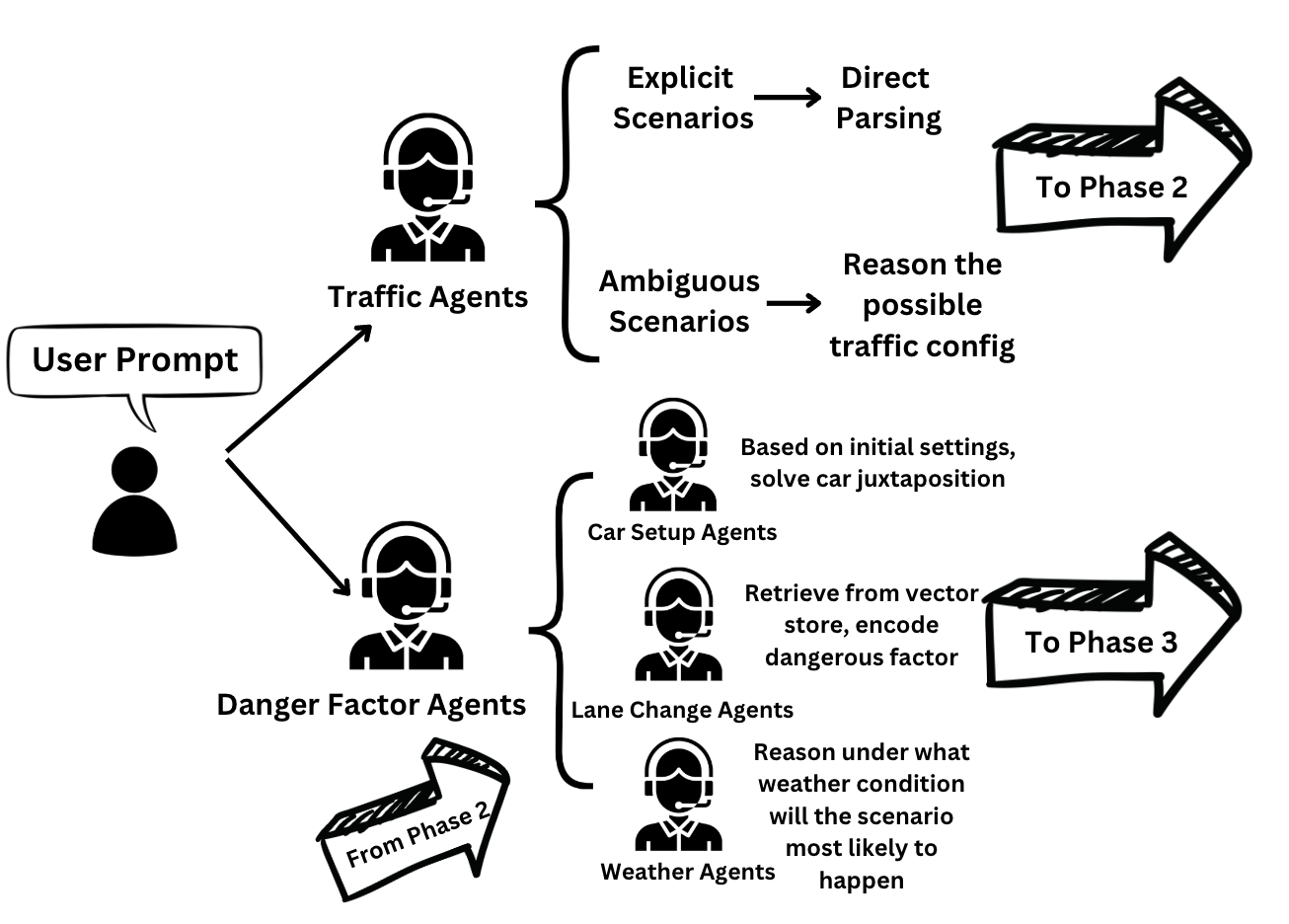}
    \caption{Agents set for the first stage to parse and reason the given description.}
    \label{fig:agents}
\end{figure}

\subsubsection{Stage 2: ASP Solver + Interactive frontend}

\begin{figure}[!ht] \centering
    \begin{tabular}{cc} 
        \begin{subfigure}{0.48\linewidth}  \centering
            \includegraphics[width=\linewidth,trim={5 5 5 5},clip]{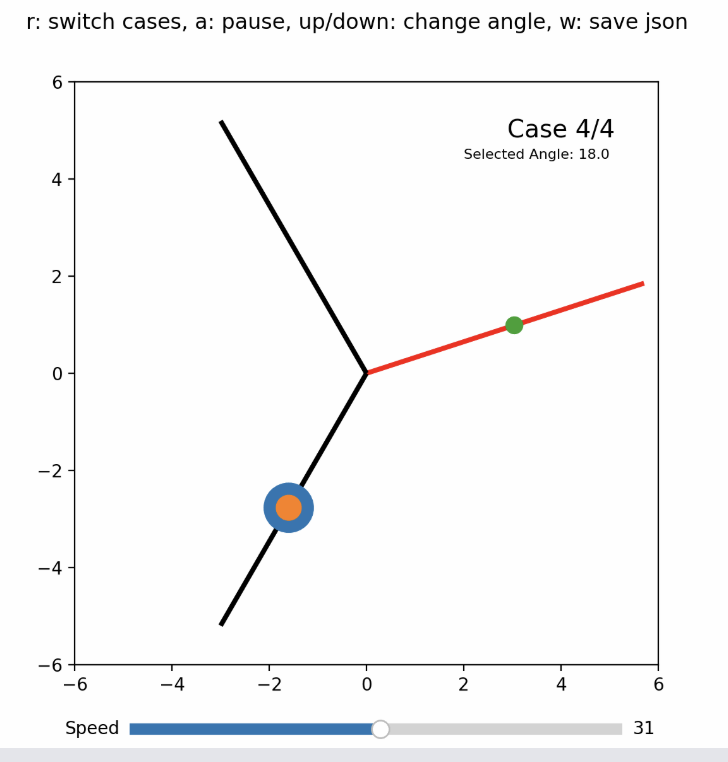}
            \caption{Logical scenario with 3 entry points and 3 cars. In total 4 cases.}
        \end{subfigure} &
        \begin{subfigure}{0.48\linewidth} \centering
            \includegraphics[width=\linewidth]{figures/traffic_3_3_mod.png}
            \caption{Logical scenario with 3 entry points and 3 cars. Case 4 with the modified angle of the road.}
        \end{subfigure} \\
        \begin{subfigure}{0.48\linewidth} \centering
            \includegraphics[width=\linewidth]{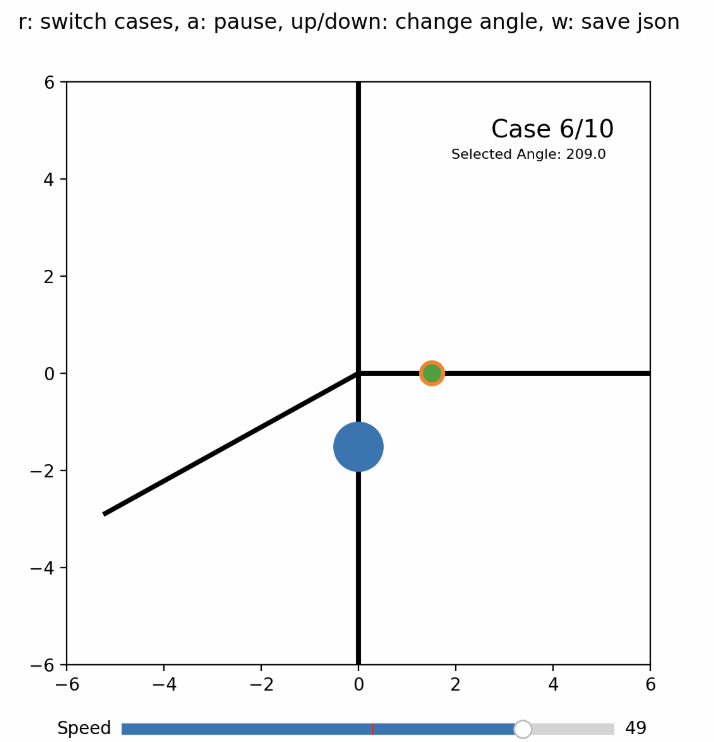}
            \caption{Logical scenario with 4 entry points and 3 cars with modified speed of cars. In total 10 cases.}
        \end{subfigure} &
        \begin{subfigure}{0.48\linewidth} \centering
            \includegraphics[width=\linewidth]{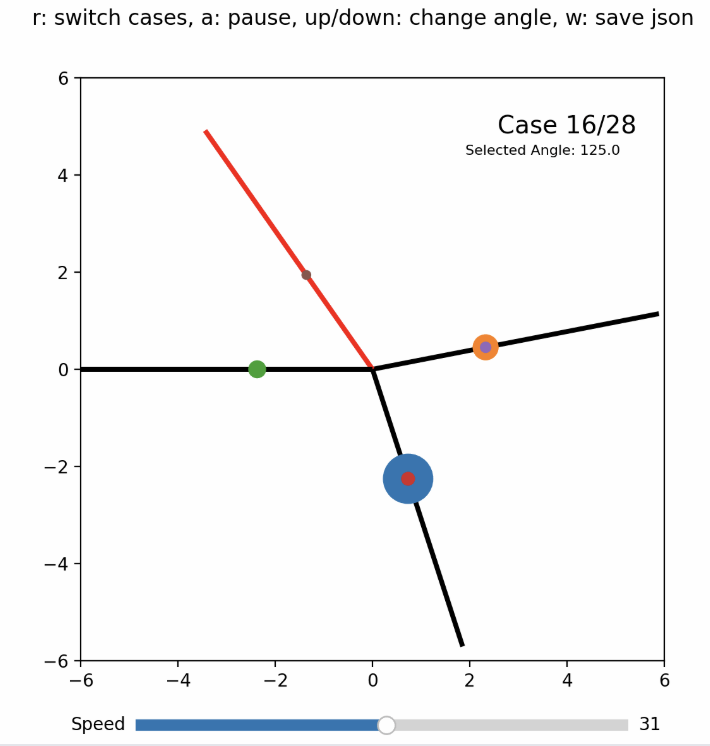}
            \caption{Logical scenario with 4 entry points and 6 cars. A more complicated scenario.}
        \end{subfigure}
    \end{tabular}
    \caption{Examples of Logical Scenarios}
    \label{fig:traffic_grid}
\end{figure}

In this project, we focus on generating all possible scenarios in uncontrolled intersections, where we can formalize the input for the ASP solver with cars, the number of entry points, and the entry point for each car.
$$
\text{car}(c), \text{entry point}(e), \text{num entries}(n)
$$

Therefore, for each car $c$ at its entry point $e$, it will have in total $n - 1$ options to enter the next $1, 2, 3,...,n-1$ road segment. Then the final exit point for the car $c$ will be 
$$
(e + 1)\ mod\ n, (e + 2)\ mod\ n, ..., (e + n -1)\ mod\ n
$$

Therefore, there will be in total of $(n-1)^c$ scenarios if we intend to calculate in this way. Therefore, we need to remove redundant scenarios. 

For the logical scenario we intend to generate, we regard every road segment as equal, not considering its real angle with another road segment and its number of road lanes. 
Therefore, the scenario as follows can be regarded as symmetric as the second traffic can be simply regarded as rotating the first traffic. 

\begin{figure}[!ht]  \centering
    \includegraphics[width=0.9\linewidth,trim={10 30 10 10},clip]{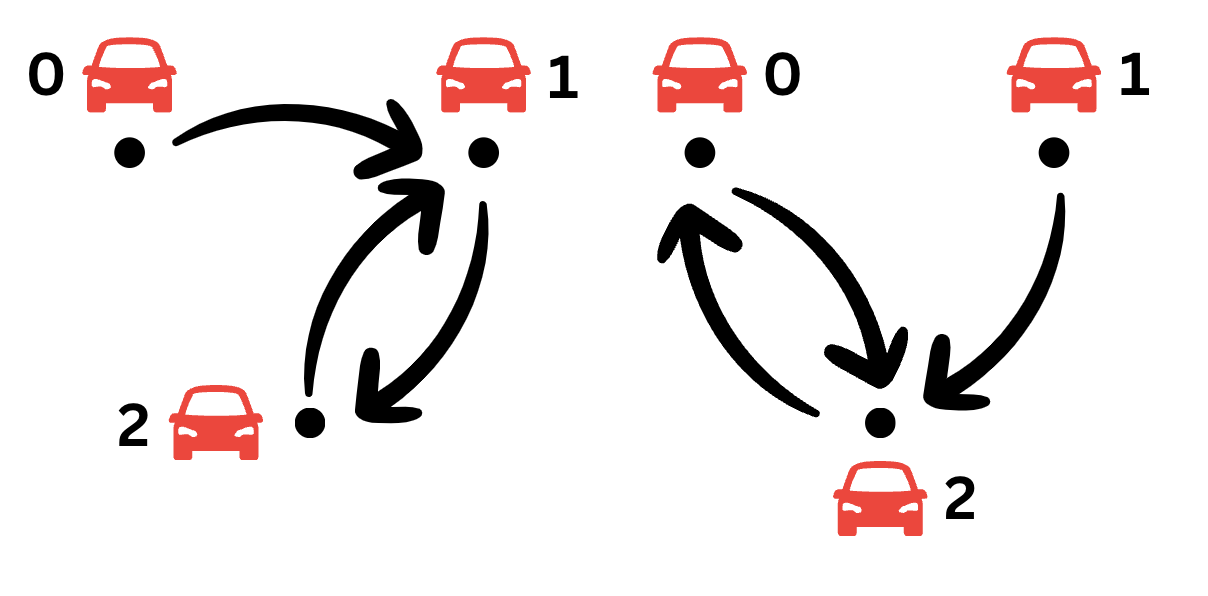}
    \caption{Symmetric cases for road traffic flow \cite{simbase}}
    \label{fig:enter-label}
\end{figure}

There is also a mathematical representation to feature this symmetry. For a symmetric situation, we can see from the left picture that car $0$ chooses to move $1$ step from its entry point, car $1$ chooses to move $1$ step from its entry point, car $2$ chooses to move $2$ step from its entry point. The symbolic movement can be represented as $(1, 1, 2)$ and the exit point can be represented as $(1, 2, 1)$. Using the same way, we can encode the right scenario to symbolic movement as $(2, 1, 1)$. \cite{simbase}

We can then notice that for a symmetric situation, the order of the symbolic movement is less important than its constituent. Therefore, scenarios that can be encoded as $(1, 1, 2), (1, 2, 1), (2, 1, 1)$ can be reduced to only one scenario. 

For a given uncontrolled intersection with 3 cars and 3 entry points, the total scenarios can be reduced from $2^3 = 8$ to only $4$ cases. Using this method, we can get comprehensive scenarios with less representation of the traffic. 

In \texttt{clingo}, we assign facts using the following semantics:

\begin{lstlisting}[language=Python]
num_entry(N)
car_entry(C, E)
symbolic_direction(1..N-1)
\end{lstlisting}

The rules are formularized as:

\begin{lstlisting}[language=Python]
% Define a symbolic move
1 { symbolic_move(C, E, D) : symbolic_direction(D) } 1 :- car_entry(C, E).

% Final move
final_move(C, E, FD) :- symbolic_move(C, E, D), num_entry(N), FD = (E + D) \ N.
\end{lstlisting}

The constraints are formularized as:

\begin{lstlisting}[language=Python]
% Ensure that each car makes exactly one symbolic move
:- symbolic_move(C, E, D1), symbolic_move(C, E, D2), D1 != D2.
\end{lstlisting}

Then we set up functional semantics to help calculate the symbolic movement patterns, which can be later used for reduction:

\begin{lstlisting}[language=Python]
% Set the count
count(X, Y) :- symbolic_direction(X), Y = #count { C : symbolic_move(C, _, X) }.
\end{lstlisting}

We have used \texttt{clorm} packaged in Python to help integrate the calling program in Python. By using \texttt{clingo}, we can set up an interface for integrated intended formularization of the traffic we intend to generate. (For example, we can further add traffic rules to form scenarios under rule-guided intersections.)

Also, to add more controllability to the logical scenarios, we designed a basic frontend to interact with the users so they can decide on which logical scenarios they prefer better. 

\subsubsection{Stage 3: Scenario Generator with LLM}

After generating the logical scenario, We have generated a set of parameters that represent a randomly generated scenario. Since it's tough for people to directly modify an OpenScenario or OpenDrive file to increase the critical level of the concrete scenario, however, the OpenScenario and OpenDrive files generated by LLM are sometimes unexpected using non-existent interfaces, we choose to generate the files with \texttt{scenariogeneration}, improving the output's stability and could customize a map for each scenario instead of reuse a map for thousands of time which restrict the variation of scenarios. We use LLM to modify initial parameters, enhancing the criticality of the specific scenario. In this stage, we use \texttt{deep-seek-chat} as the model to update the parameters. Considering the unstable output of the LLM, we have implemented a script to verify the validity of the final parameters. If any parameters are deemed invalid, we randomly assign values to them.

The parameters used in the config files are shown in \autoref{tab:param}.

\begin{table}[!ht]  \centering \small
    \begin{tabular}{|p{2.6cm}|p{5cm}|} \hline
        \textbf{Category} & \textbf{Details} \\  \hline
        \multicolumn{2}{|l|}{\textbf{Roads}} \\   \hline
        \textbf{road\_id} & Identifier for the road. \\  \hline
        \textbf{road\_len} & Length of the road. \\  \hline
        \textbf{angle} & Angle between this road and the previous road. For the first road, it is the angle between it and the original direction. \\  \hline
        \textbf{left\_num} & Number of left lanes. \\  \hline
        \textbf{right\_num} & Number of right lanes. \\ \hline
        \multicolumn{2}{|l|}{\textbf{Cars}} \\  \hline
        \textbf{name} & Name of the car. \\  \hline
        \textbf{type} & Type of the car. \\   \hline
        \textbf{init\_pos} & Initial position of the car. \\  \hline
        \textbf{init\_speed} & Initial speed of the car. \\  \hline
        \textbf{init\_road\_id} & Road where the car initially is. \\ \hline
        \textbf{init\_lane\_id} & Lane where the car initially is. \\  \hline
        \textbf{turning\_pos} & Position where the car starts to veer. \\ \hline
        \textbf{final\_pos} & Position where the car finishes veering. \\ \hline
        \textbf{final\_road\_id} & Final road where the car is. \\ \hline
        \textbf{final\_lane\_id} & Final lane where the car is. \\ \hline
        \multicolumn{2}{|l|}{\textbf{Change Lanes}} \\ \hline
        \textbf{car\_name} & Name of the car. \\  \hline
        \textbf{change\_lane\_pos} & Distance traveled by the car before it starts to change lanes. \\  \hline
        \textbf{lane\_id\_after\_change} & Final lane ID after the car changes lanes. \\  \hline
    \end{tabular}
    \caption{Parameters specified as input for the LLMs.}
    \label{tab:param}
\end{table}


For better adapting to our goals, we propose a three-stages framework to generate comprehensive scenarios combining critical factors. This framework integrates ASP solver and LLM, which provides the framework with scalability to adapt to more complicated scenario generation. 

\section{EXPERIMENTAL SETUP}
\label{sec:experiments}

\subsection{Esmini Installation}

Refer to \href{https://github.com/esmini/esmini}{Esmini official GitHub repo} to find installation steps for setting up the Esmini environment. We have used version \texttt{2.40.1} for our project. 

\subsection{Software and Hardware environment}

You can check out the repo and use \texttt{pip install -r requirements.txt} to load the environment, the essential package we use in our environment is included in \autoref{tab:soft} and the hardware configuration is included in \autoref{tab:hard}.

\begin{table}[!ht] \centering
    \begin{tabular}{|c|c|} \toprule
    \multicolumn{2}{|c|}{\textbf{Software Configuration}} \\ \hline
    \multicolumn{2}{|c|}{For Stage 1 and Stage 3} \\   \hline
    \textbf{Package} & \textbf{Version} \\ \hline
    python & 3.9.19 \\
    pip & 24.0 \\
    langchain & 0.2.4 \\
    langchain-chroma & 0.1.1 \\
    langchain-community & 0.2.4 \\
    langchain-core & 0.2.6 \\
    langchain-openai & 0.1.8 \\
    openai & 1.34.0 \\ \hline
    \multicolumn{2}{|c|}{For Stage 2} \\  \hline
    \textbf{Package} & \textbf{Version} \\ \hline
    clingo & 5.7.1 \\
    Clorm & 1.6.0 \\
    matplotlib & 3.9.0 \\
    numpy & 1.26.4 \\   \hline
    \multicolumn{2}{|c|}{For Stage 3} \\    \hline
    \textbf{Package} & \textbf{Version} \\     \hline
    pygame & 2.5.2 \\
    scenariogeneration & 0.14.9 \\    \hline
    \end{tabular}
    \caption{Software Configuration}
    \label{tab:soft}
\end{table}

\begin{table}[!ht]  \centering
    \begin{tabular}{|l|c|} \toprule
    \multicolumn{2}{|c|}{\textbf{CPU Configuration}} \\  \hline
    \textbf{Model} & Intel(R) Core(TM) i9-14900KF \\  \hline
    \textbf{CPUs} & 32 \\  \hline
    \textbf{CPU GHz} & 3.2 \\ \hline
    \textbf{CPU Max GHz} & 6 \\ \hline
    \textbf{L1d cache} & 576 KiB \\  \hline
    \textbf{L1i cache} & 384 KiB \\ \hline
    \textbf{L2 cache} & 24 MiB \\ \hline
    \textbf{Memory} & 32 GiB \\  \hline
    \textbf{Swap Area} & 62.5 GiB \\  \hline
    \textbf{Operating System} & Ubuntu 20.04.6 LTS (Focal) \\ \hline
    \end{tabular}
    \caption{CPU Configuration}
    \label{tab:cpu}
\end{table}

\begin{table}[!ht]  \centering
    \begin{tabular}{|c|c|} \hline
    \multicolumn{2}{|c|}{\textbf{GPU Configuration}} \\  \hline
    \textbf{Model} & NVIDIA GeForce RTX 4090 \\ \hline
    \textbf{Memory} & 24 GiB \\ \hline
    \end{tabular}
    \caption{Hardware Configuration}
    \label{tab:hard}
\end{table}

We are testing our framework on the following three categories. 

\subsection{Intersection Settings}
\begin{enumerate}
    \item \textbf{Four-Way Intersection - High Traffic Flow}: 
    \emph{"A four-way intersection with vehicles approaching from all directions at varying speeds, including one vehicle attempting an unprotected left turn."}
    
    \item \textbf{T-Intersection - Limited Visibility}: 
    \emph{"A T-intersection with obstructed views due to parked cars and a speeding vehicle approaching the main road while another vehicle tries to merge."}
    
    \item \textbf{Roundabout with Multiple Exits}: 
    \emph{"A roundabout with multiple exits, where one vehicle changes lanes abruptly to exit, causing potential conflict with another vehicle entering at high speed."}
    
    \item \textbf{Skewed Intersection - Sharp Angles}: 
    \emph{"A skewed intersection where vehicles approach at non-right angles, including a fast-moving car trying to overtake another vehicle while crossing."}
    
    \item \textbf{Y-Intersection}: 
    \emph{"A Y-intersection where a pedestrian suddenly crosses while a vehicle speeds to make a left turn, and another vehicle merges from the right."}
\end{enumerate}

\subsection{Dangerous Factors}
\begin{enumerate}
    \item \textbf{Sudden Lane Changes}: 
    \emph{"A car abruptly changes lanes in a roundabout while another vehicle accelerates to exit, causing a near-collision scenario."}
    
    \item \textbf{Speeding Vehicles}: 
    \emph{"Two speeding vehicles approach a T-intersection simultaneously from opposite directions, one trying to make an unprotected left turn."}
    
    \item \textbf{Tailgating}: 
    \emph{"A vehicle tailgating another car at a four-way stop, leading to a sudden braking event as a pedestrian steps into the crosswalk."}
    
    \item \textbf{Delayed Braking Response}: 
    \emph{"A distracted driver delays braking at a Y-intersection while another car merges unexpectedly from the right."}
    
    \item \textbf{Merging Conflicts}: 
    \emph{"A car aggressively merges into traffic at high speed in a roundabout while another vehicle prepares to exit."}
    
    \item \textbf{Multiple Cars Scenarios}: 
    \emph{"Ten cars arriving at a T intersection."}
\end{enumerate}

\section{RESULTS}
\subsection{LLM analyzing ability}
LLM demonstrates good capability of understanding the description of a scenario in natural language. In phase 1, LLM could extract information on the proper number of cars and entrances described both ambiguously and explicitly. In our testing, it has 100\% correctness (for ambiguous descriptions, generating proper values is considered correct). 

\begin{table}[!ht] \centering
    \begin{tabular}{ccccc} \toprule
        \textbf{Case} & \textbf{Num of Cars} & \textbf{Num of Entries} & \textbf{Initial Position} & \textbf{T/F} \\ \hline
            $a.1$ & 4 & 4 & [0,1,2,3] & T \\
            $a.2$ & 3 & 3 & [0,1,2] & F \\
            $a.3$ & 2 & 4 & [0,3] & T  \\
            $a.4$ & 2 & 4 & [1,3] & T \\
            $a.5$ & 2 & 3 & [0,1] & T \\
            $b.1$ & 2 & 4 & [1,3] & T \\
            $b.2$ & 2 & 3 & [0,1] & T  \\
            $b.3$ & 2 & 4 & [1,3] & F \\
            $b.4$ & 2 & 3 & [0,1] & T \\
            $b.5$ & 2 & 4 & [1,3] & T  \\
            $b.6$ & 10 & 3 & [0,1,2,0,1,2,0,1,2,0] & T \\ \hline
    \end{tabular}
    \caption{Results parsed/reasoned by the LLM for stage 2}
    \label{fig:result-2}
\end{table}

\begin{table}[!ht] \centering
    \begin{tabular}{l|l} \toprule
        \textbf{Case} & \textbf{Percent} \\ \hline
        Road information analyze & 11/11 \\
        \# of Cars analyze & 11/11 \\
        Not support actions & 2/11 \\ 
        Total Correctness & 9/11 \\ \hline
    \end{tabular}
    \caption{Scenario-analyze-correctness by LLMs.}
    \label{fig:scenario-analyze}
\end{table}

\subsection{LLMs modification ability}
Although LLM demonstrates a relatively good capability of coding, it is limited by the size of their training corpus and the popularity of the specific language. Such generation for underfitting tasks will result in greater illusion. We use LLM to parse hazardous factors and let it modify the parameters list we provide. 

\begin{figure}[!ht] \centering \small
    \begin{tabular}{llc}   \hline
        \textbf{Case} & \textbf{Modified Parameters} & \textbf{Correctness} \\  \hline
        $a.1$ & angle, init\_speed, change\_lane & F \\ 
        $a.2$ & angle, init\_speed, change\_lane & T \\
        $a.3$ & angle, init\_speed, change\_lane & F \\ 
        $a.4$ & angle, init\_speed, change\_lane & T \\
        $a.5$ & angle, init\_speed, change\_lane & T \\
        $b.1$ & angle, init\_speed, change\_lane & T \\
        $b.2$ & angle, init\_speed, change\_lane & F \\ 
        $b.3$ & angle, init\_speed, change\_lane & T \\
        $b.4$ & angle, init\_speed, change\_lane & T \\
        $b.5$ & angle, init\_speed, change\_lane & T \\
        $b.6$ & angle, init\_speed, change\_lane & T \\
        \hline
    \end{tabular}
    \caption{Parameters modified by LLMs}
    \label{fig:parameters-modified}
\end{figure}

\begin{table}[!ht] \centering
    \begin{tabular}{|c|c|c|c|} \toprule
    & \textbf{Logical Scenarios} & \textbf{Concrete Scenarios} & \textbf{M} \\  \hline
    $a.1$ & \includegraphics[width=0.2\linewidth,trim={130 130 130 130},clip]{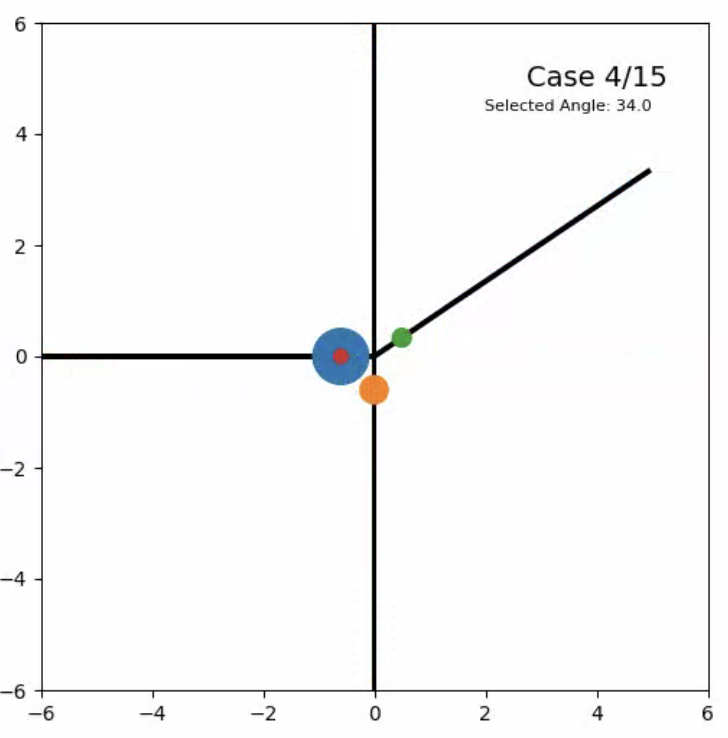} & \includegraphics[width=0.3\linewidth]{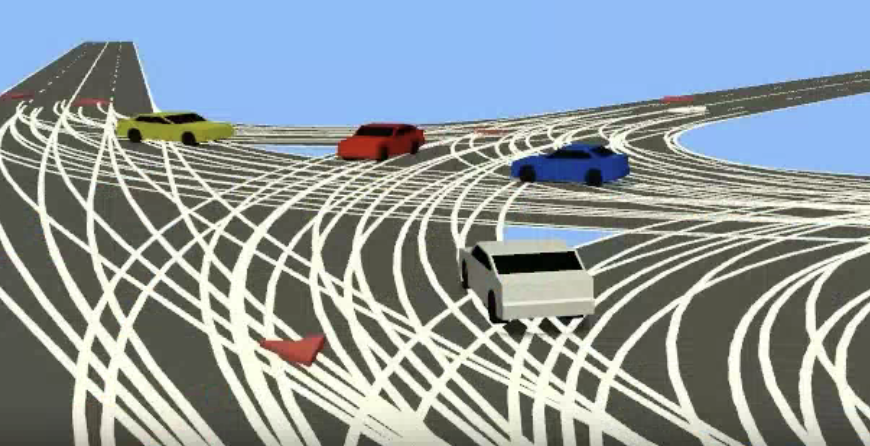} & T \\  \hline
    $a.2$ & \includegraphics[width=0.2\linewidth,trim={130 130 130 130},clip]{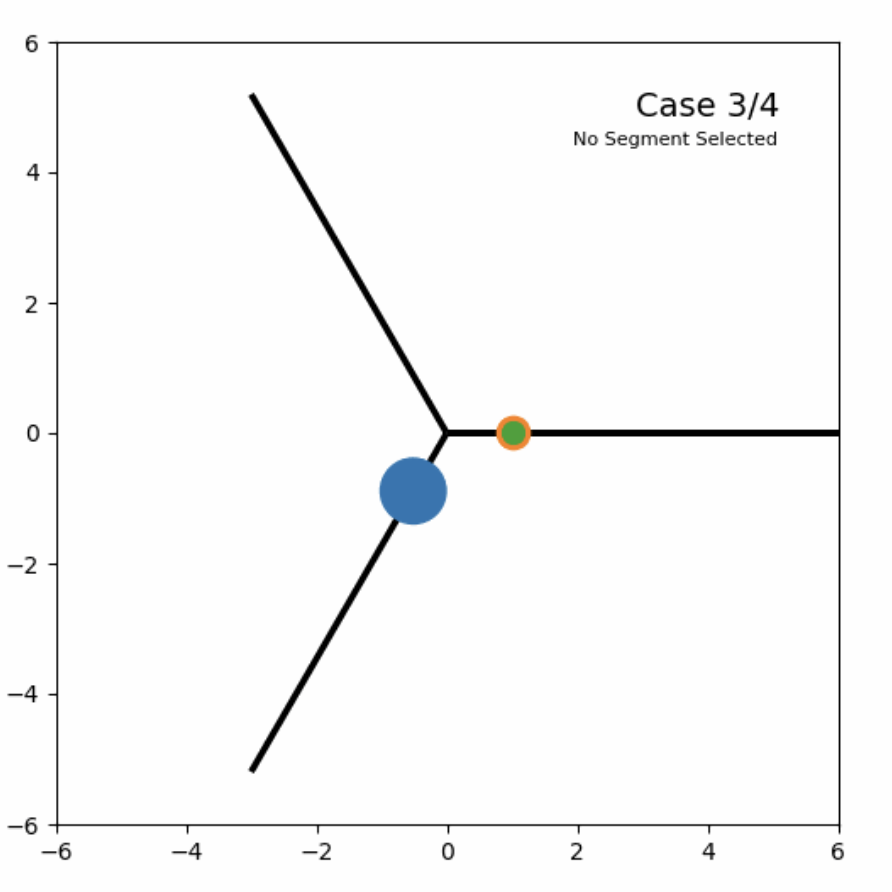} & \includegraphics[width=0.3\linewidth]{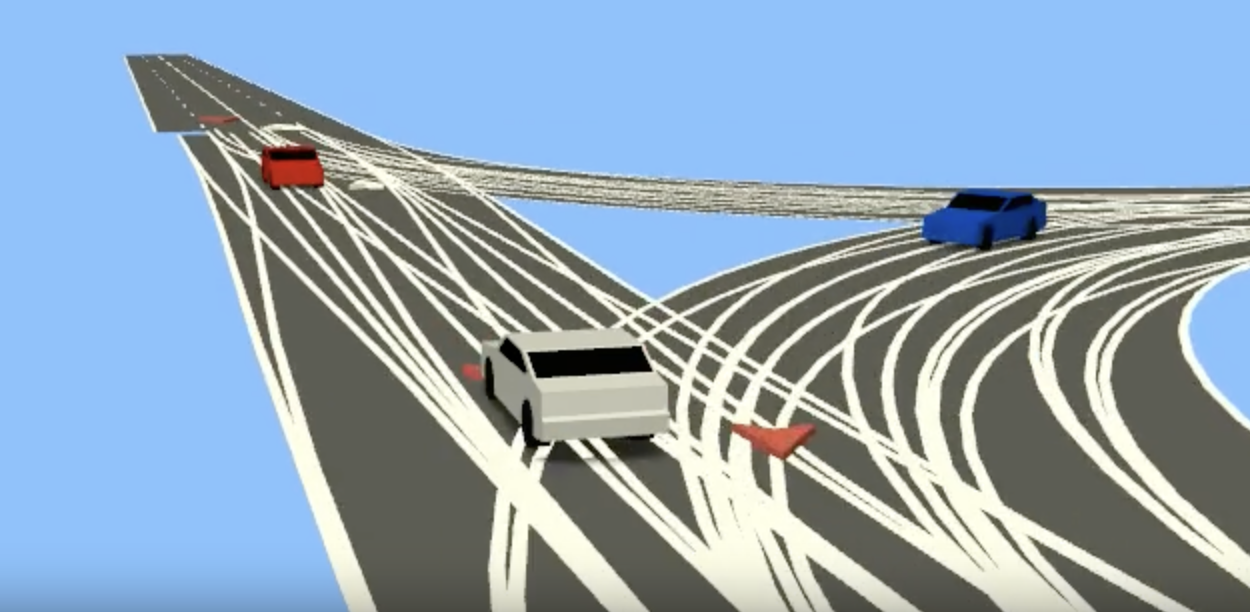} & T \\  \hline
    $a.3$ & \includegraphics[width=0.15\linewidth,trim={130 130 130 130},clip]{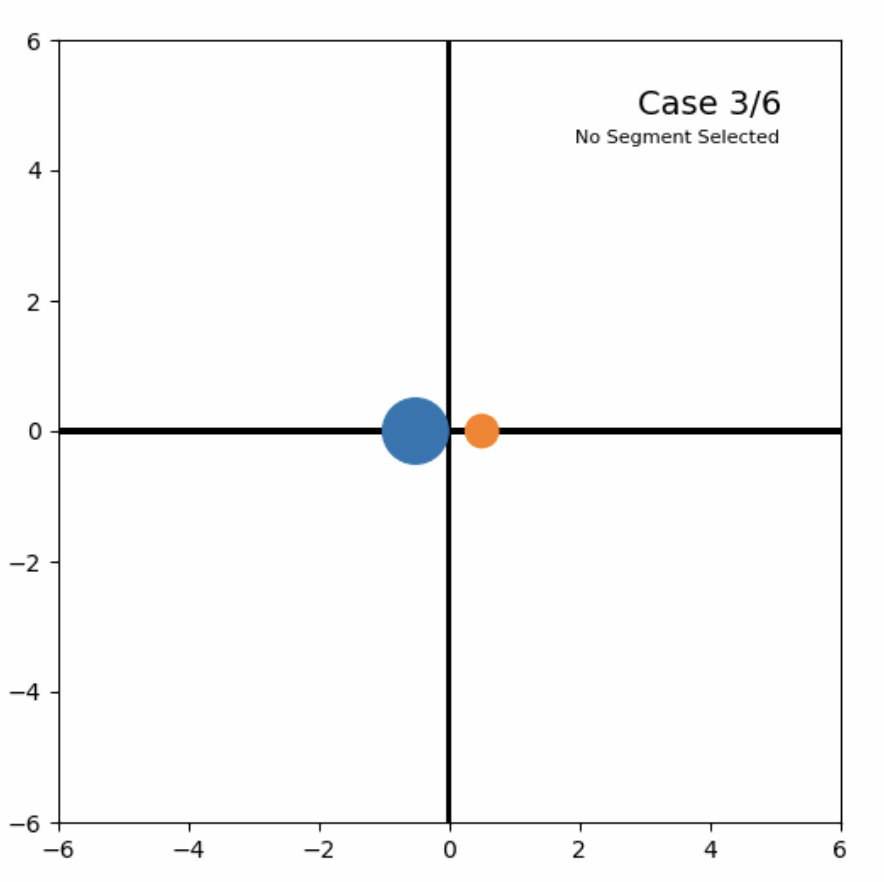} & \includegraphics[width=0.3\linewidth]{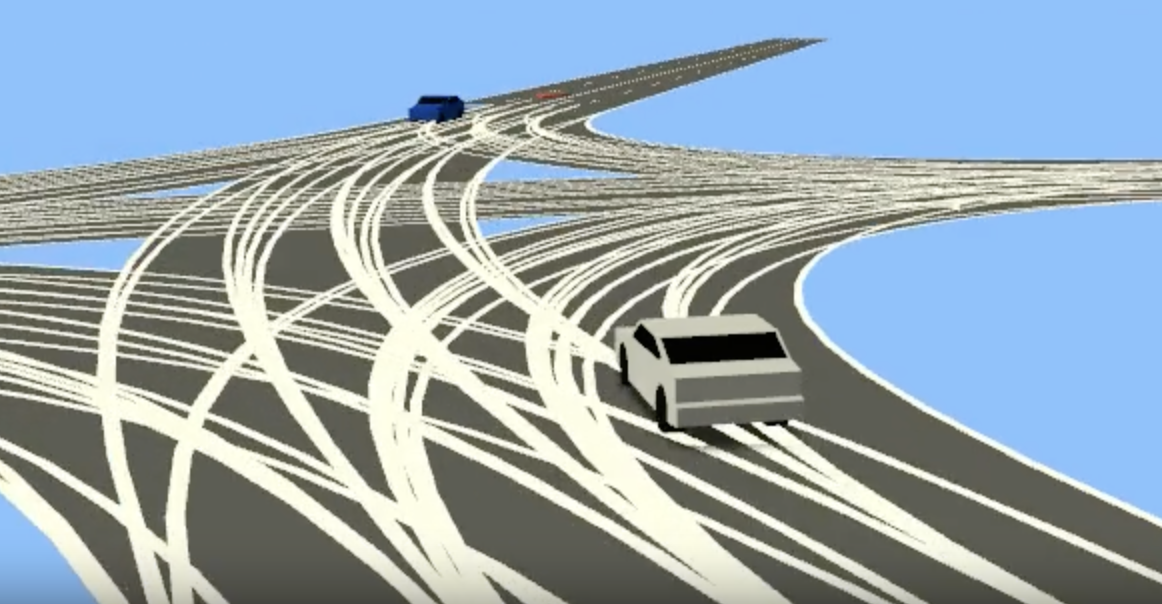} & T \\  \hline
    $a.4$ & \includegraphics[width=0.15\linewidth,trim={160 140 100 140},clip]{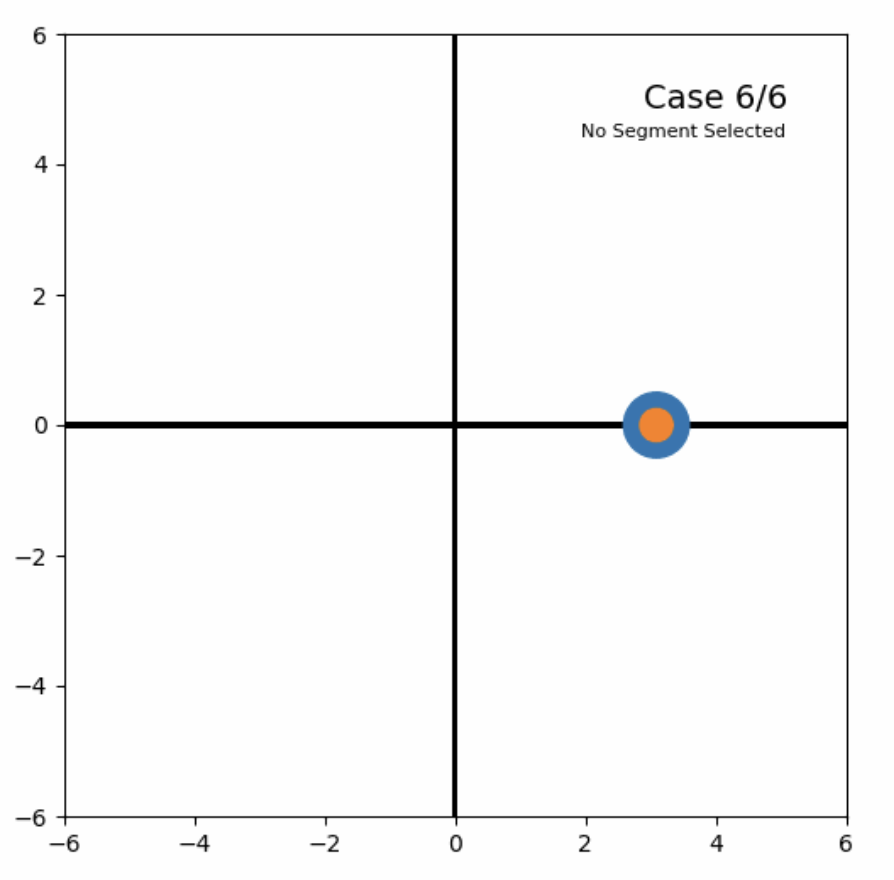} & \includegraphics[width=0.3\linewidth]{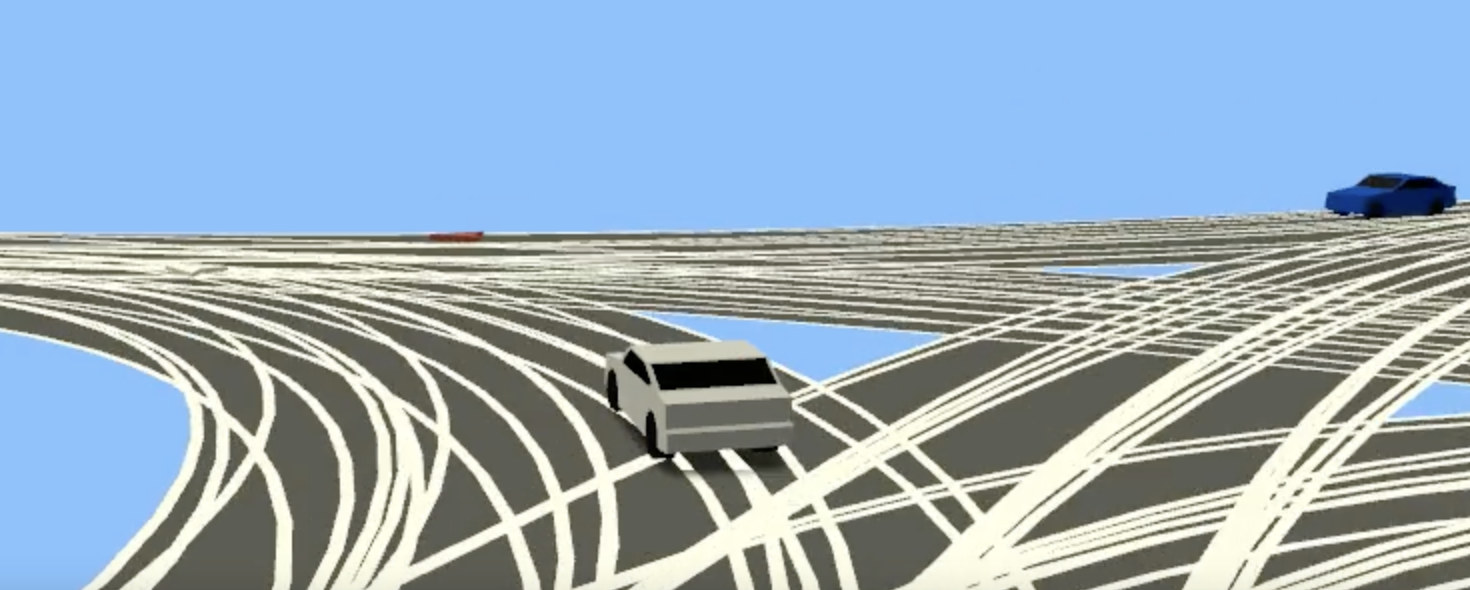} & T \\  \hline
    $a.5$ & \includegraphics[width=0.15\linewidth,trim={130 130 130 120},clip]{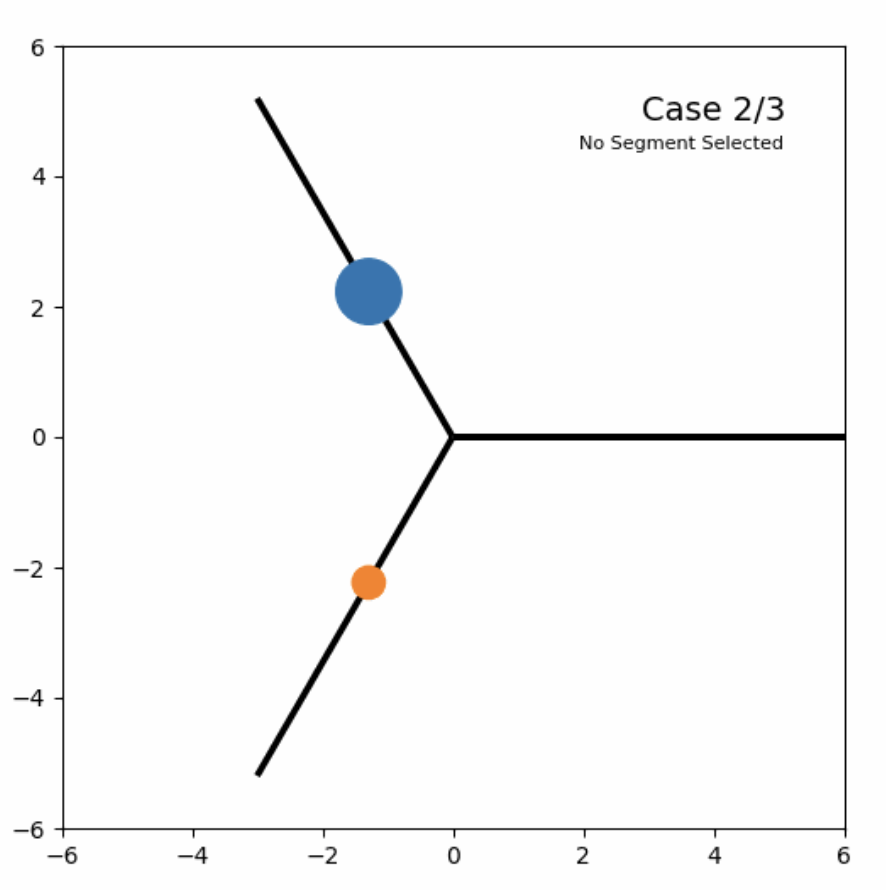} & \includegraphics[width=0.3\linewidth]{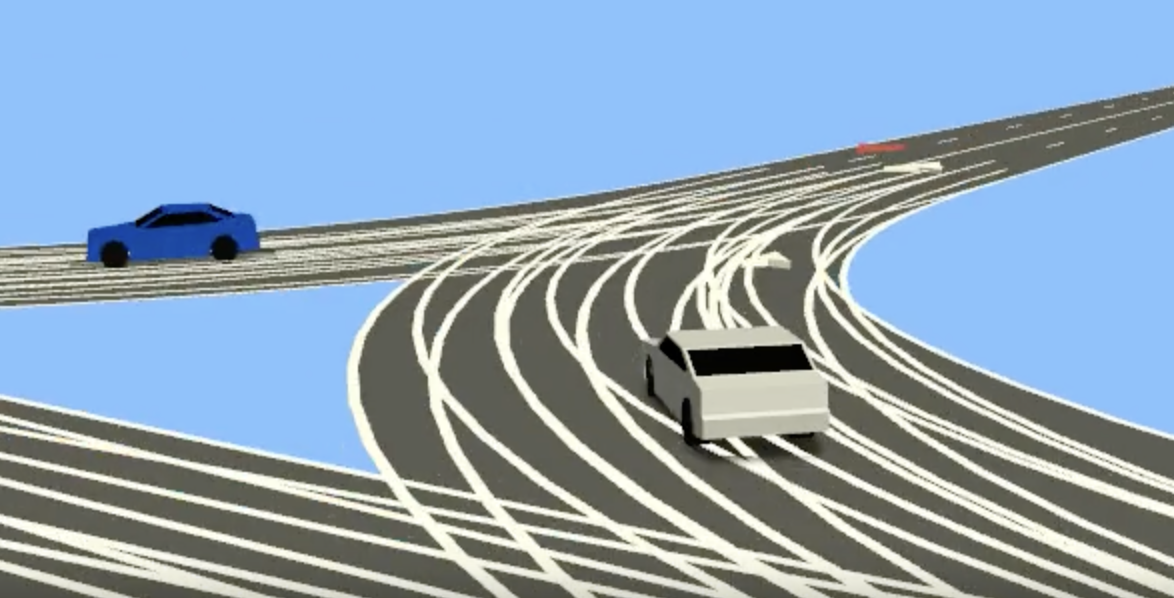} & T \\  \hline
    $b.1$ & \includegraphics[width=0.15\linewidth,trim={130 130 130 130},clip]{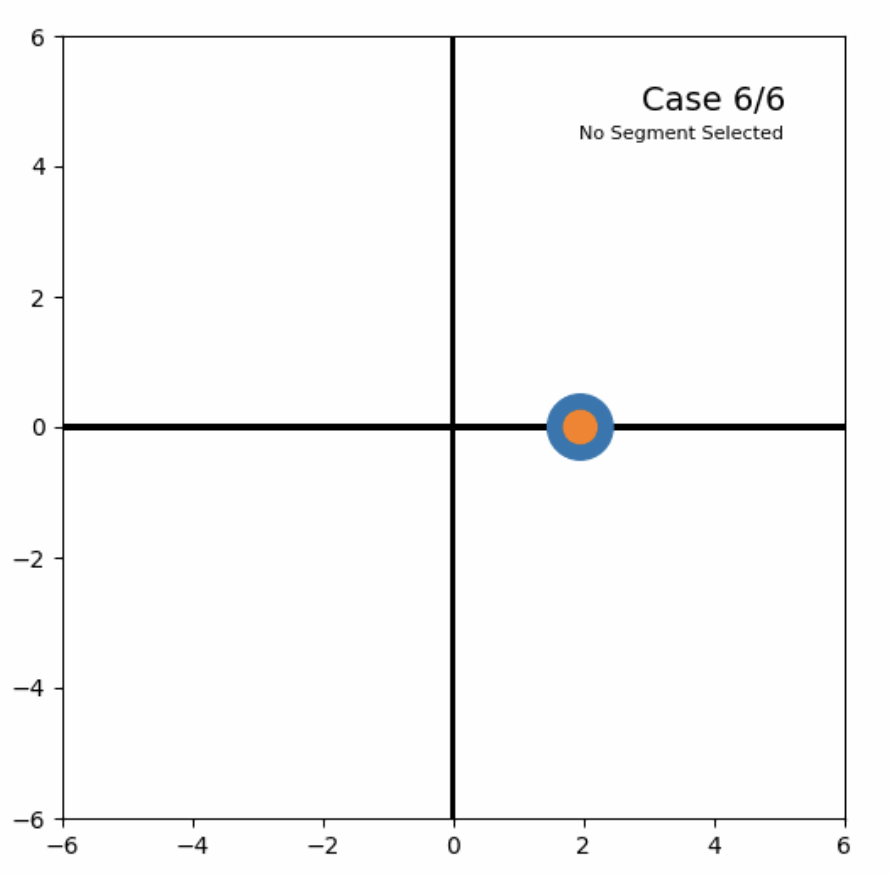} & \includegraphics[width=0.3\linewidth]{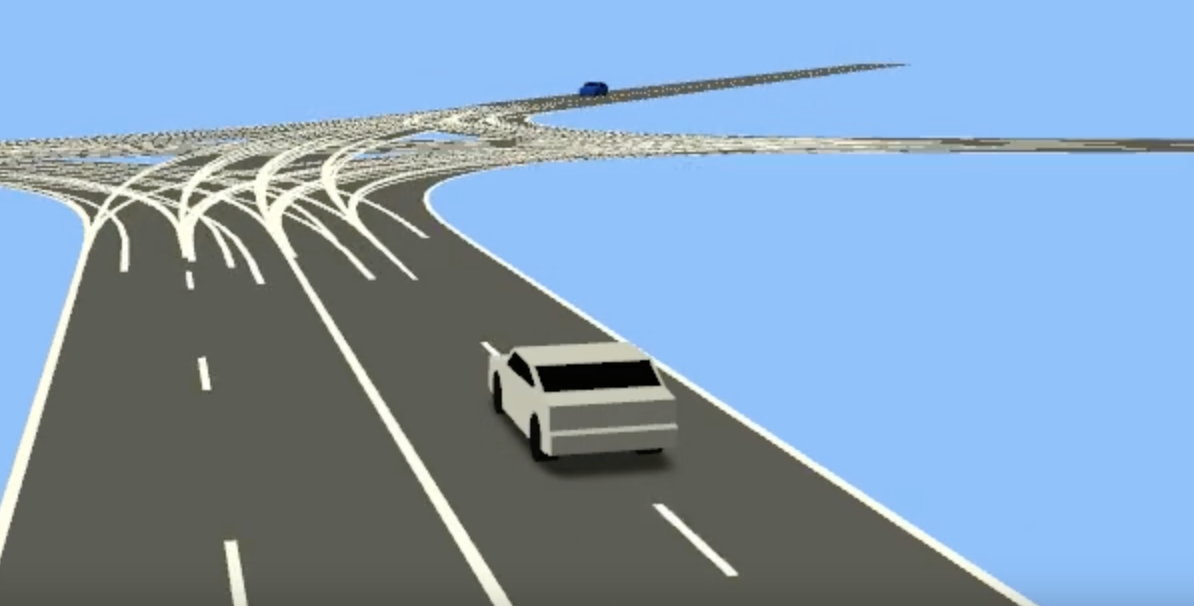} & T \\  \hline
    $b.2$ & \includegraphics[width=0.15\linewidth,trim={90 90 90 90},clip]{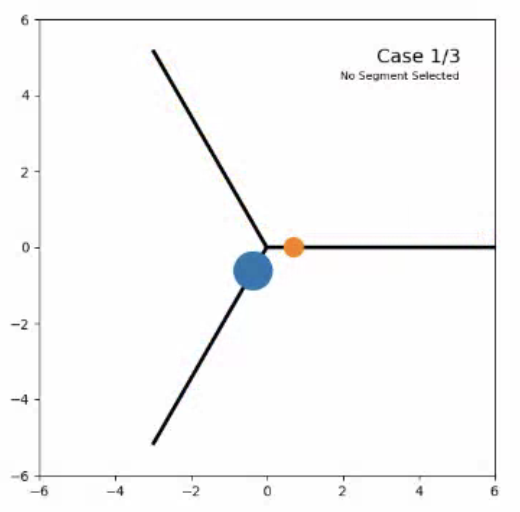} & \includegraphics[width=0.3\linewidth]{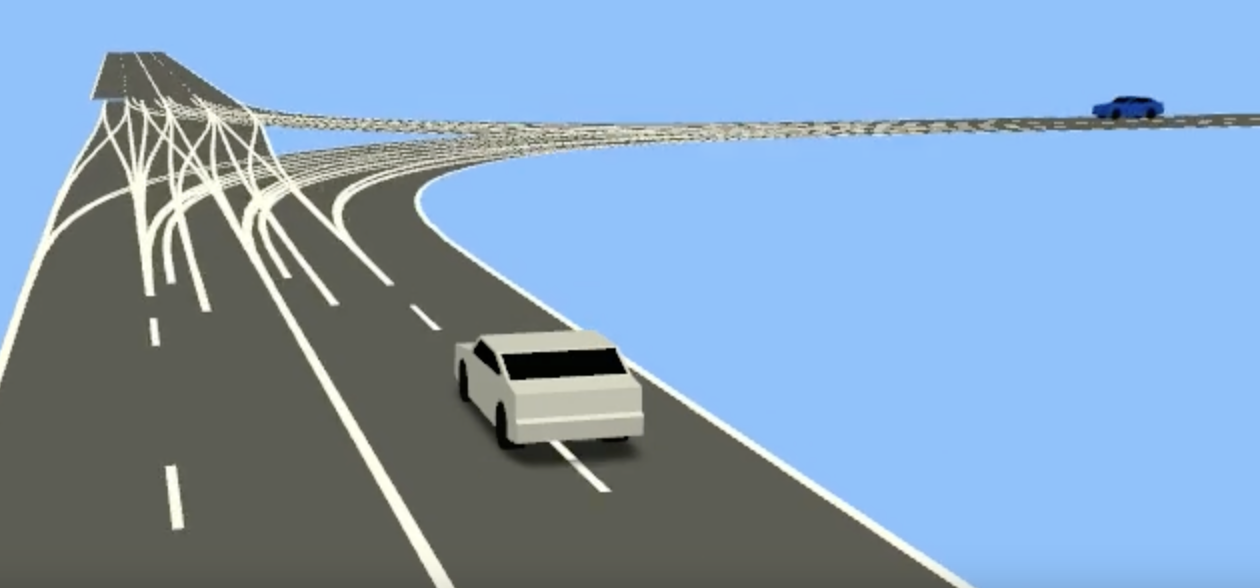} & T \\  \hline
    $b.3$ & \includegraphics[width=0.15\linewidth,trim={110 110 110 110},clip]{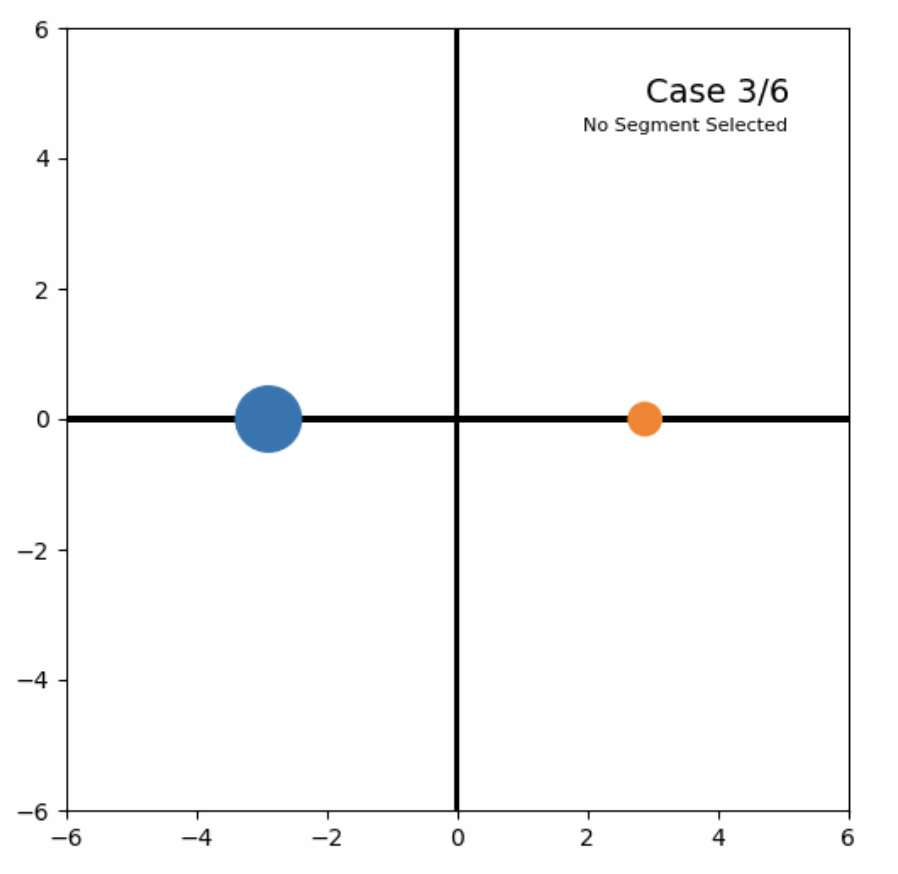} & \includegraphics[width=0.3\linewidth]{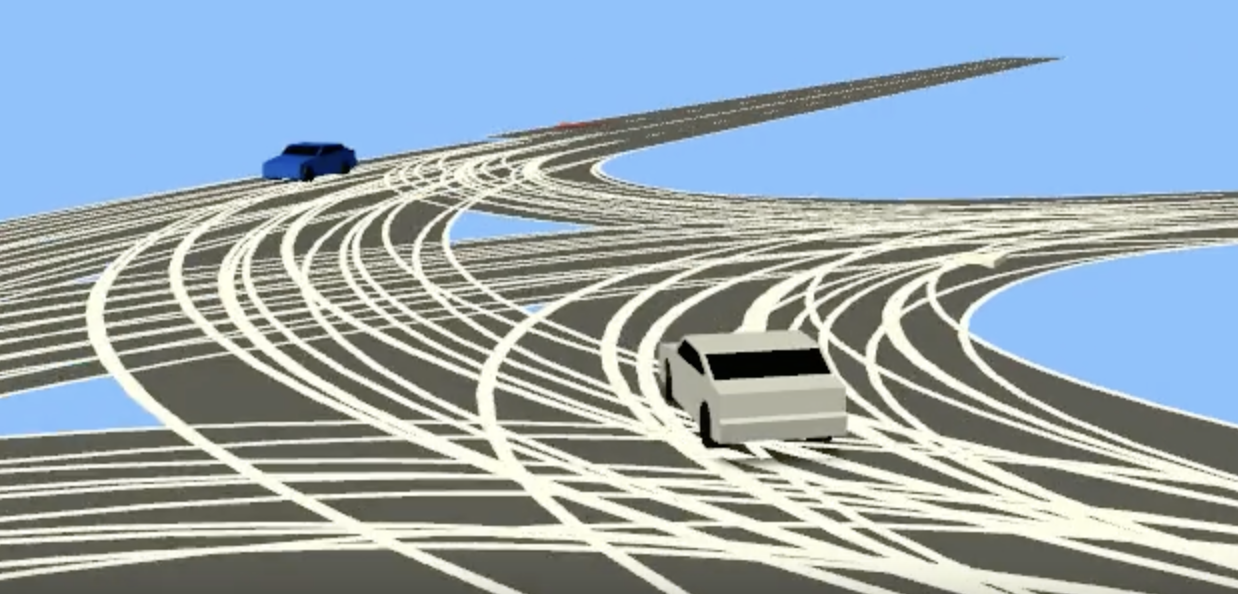} & T \\  \hline
    $b.4$ & \includegraphics[width=0.15\linewidth,trim={110 110 150 150},clip]{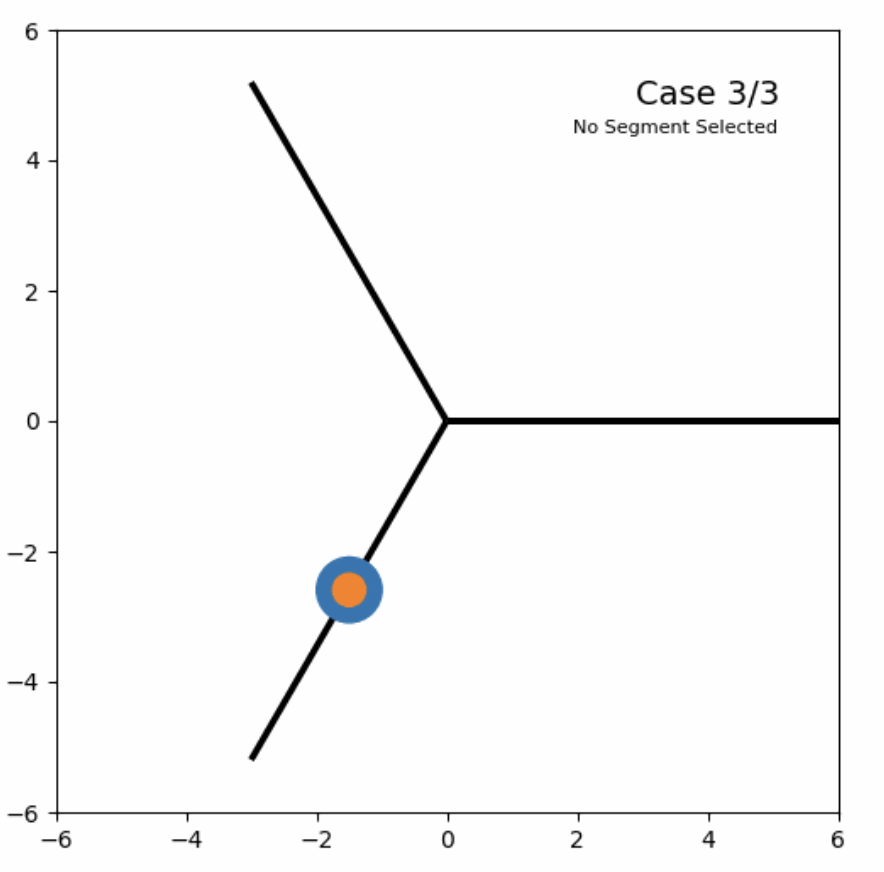} & \includegraphics[width=0.3\linewidth]{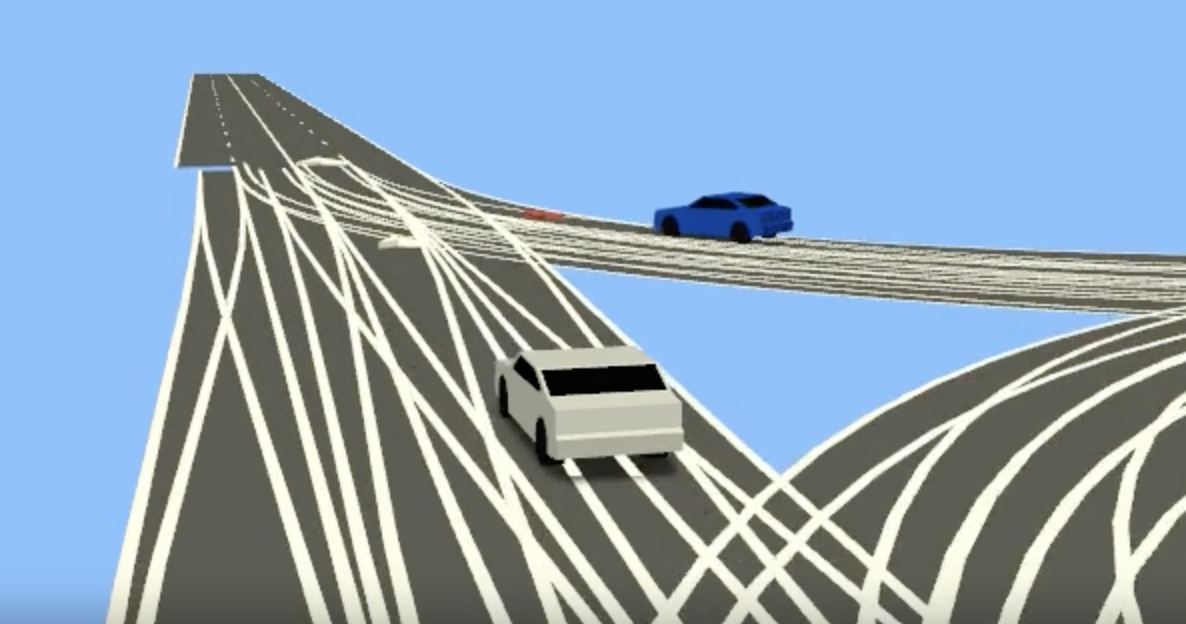} & T \\  \hline
    $b.5$ & \includegraphics[width=0.15\linewidth,trim={130 130 130 130},clip]{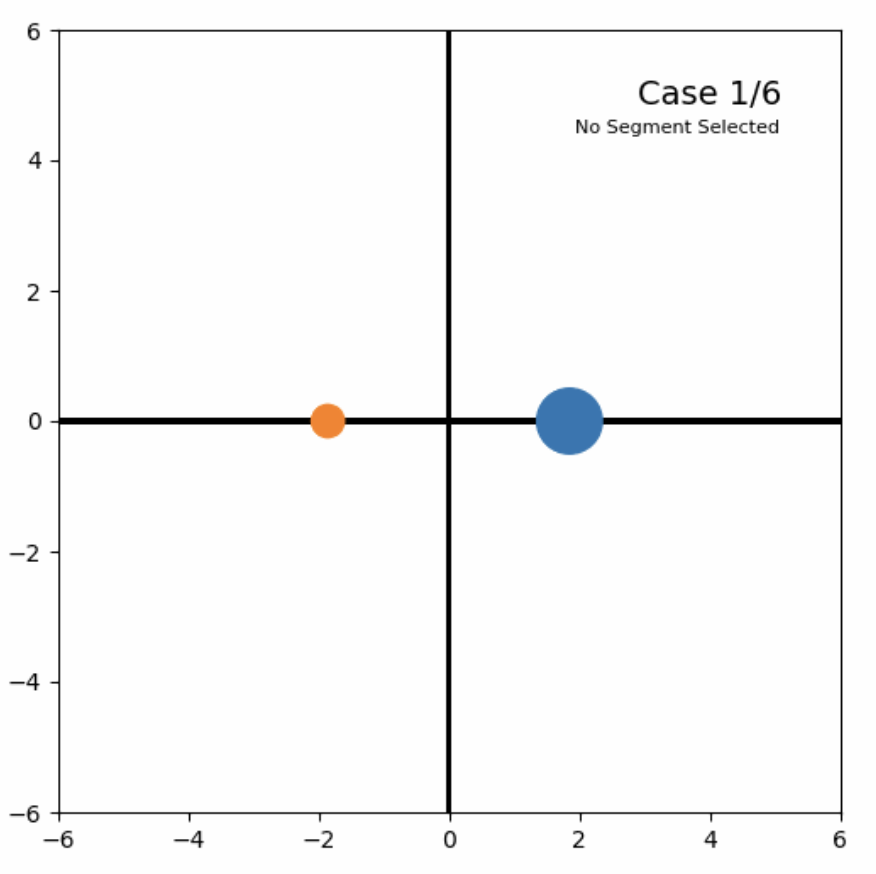} & \includegraphics[width=0.3\linewidth]{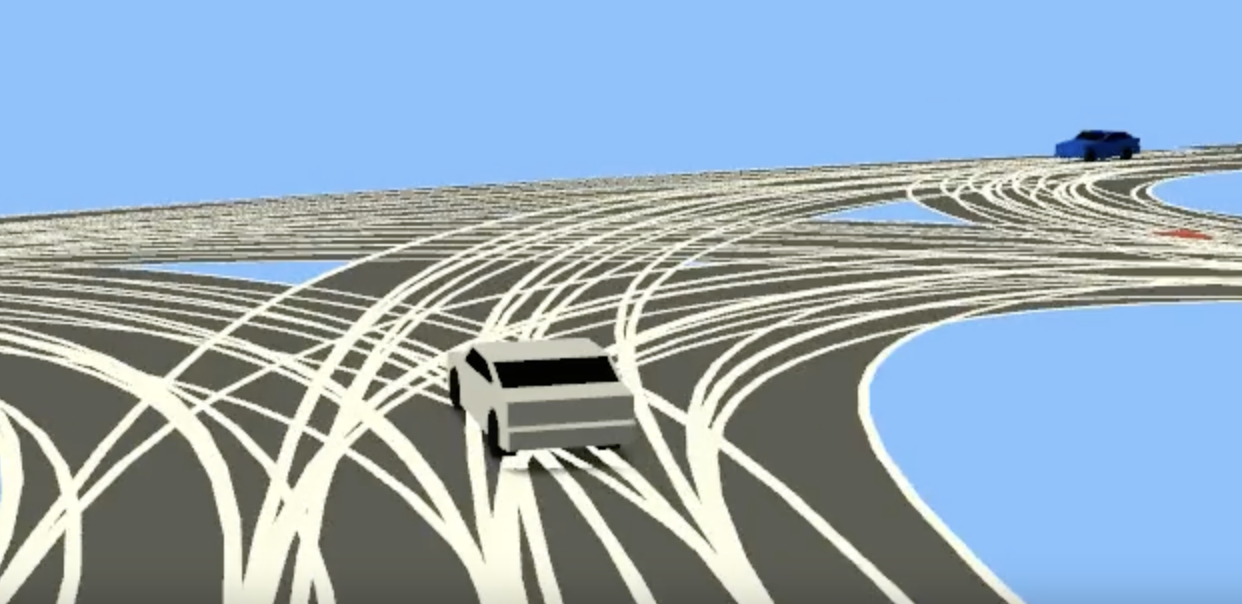} & T \\  \hline
    $b.6$ & \includegraphics[width=0.2\linewidth,trim={30 50 20 40},clip]{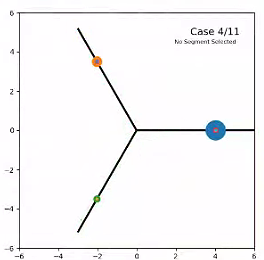} & \includegraphics[width=0.3\linewidth]{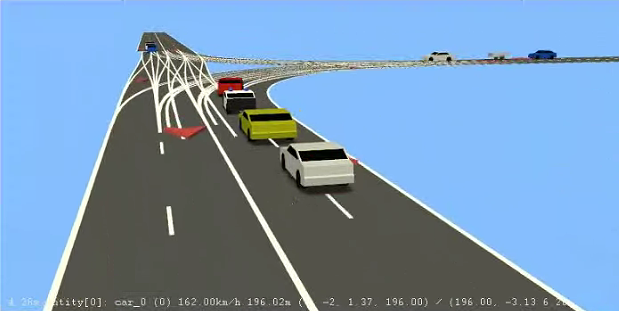} & T \\ \hline
    \end{tabular}
\end{table}

\section{CONCLUSIONS}
\label{sec:conclusions}

The ML-SceGen framework represents a significant advancement in the field of scenario generation for autonomous vehicles, particularly in managing the complexities of uncontrolled intersections. By integrating Large Language Models (LLMs) with Answer Set Programming (ASP), the framework overcomes key limitations present in existing methods, such as lack of controllability, scalability, and realism. The three-stage process—transitioning from Functional Scenarios to Logical Scenarios and finally to Concrete Scenarios—ensures a comprehensive representation of diverse driving conditions that include critical danger factors. This enhanced approach not only allows users to harness controllability over generated scenarios but also addresses the urgent need for comprehensive scenarios that reflect real-world complexities and risks. 
The promising results illustrate the potential of this framework to contribute greatly to the training of safer and more reliable autonomous driving systems.

In the future, we will improve this work from five aspects:
\begin{enumerate}
    \item Improve the construction of the ASP solver to support more types of scenarios
    \item Add more dangerous components: acceleration/deceleration, support for pedestrians and cyclers in the scenario generator.
    \item Build up more realistic scenarios inside the Carla simulator (Currently, the API has some compatibility issues to be solved.)
    \item We need to find further metrics to evaluate the generated results.
    \item Optimize the prompting process
\end{enumerate}

\bibliographystyle{IEEEtran}
\bibliography{references}

\end{document}